\documentclass[runningheads]{llncs}
\usepackage[T1]{fontenc}
\usepackage{graphicx}
\usepackage{booktabs}
\usepackage{subfigure}

\usepackage{times}
\usepackage{latexsym}
\usepackage{amsmath, mathtools}
\usepackage{makecell}
\usepackage{times}
\usepackage{latexsym}
\usepackage{graphicx}
\usepackage[T1]{fontenc}

\usepackage[utf8]{inputenc}

\usepackage{microtype}

\begin{document}

\title{Data Augmentation of Multi-turn Psychological Dialogue via Knowledge-driven Progressive Thought Prompting}
%
%
\author{Jiyue Jiang\inst{1,2} \and
Liheng Chen\inst{1} \and
Sheng Wang\inst{1} \and
Lingpeng Kong\inst{1} \and
Yu Li\inst{2} \and
Chuan Wu\inst{1}}
\authorrunning{J. Author et al.}
%
\institute{The University of Hong Kong \and
The Chinese University of Hong Kong }
\maketitle              
\begin{abstract}

Existing dialogue data augmentation (DA) techniques predominantly focus on augmenting utterance-level dialogues, which makes it difficult to take dialogue contextual information into account. The advent of large language models (LLMs) has simplified the implementation of multi-turn dialogues. Due to absence of professional understanding and knowledge, it remains challenging to deliver satisfactory performance in low-resource domain, like psychological dialogue dialogue. DA involves creating new training or prompting data based on the existing data, which help the model better understand and generate psychology-related responses. In this paper, we aim to address the issue of multi-turn dialogue data augmentation for boosted performance in the psychology domain. We propose a knowledge-driven progressive thought prompting method to guide LLM to generate multi-turn psychology-related dialogue. This method integrates a progressive thought generator, a psychology knowledge generator, and a multi-turn dialogue generator. The thought generated by the progressive thought generator serves as a prompt to prevent the generated dialogue from having significant semantic deviations, while the psychology knowledge generator produces psychological knowledge to serve as the dialogue history for the LLM, guiding the dialogue generator to create multi-turn psychological dialogue. To ensure the precision of multi-turn psychological dialogue generation by LLM, a meticulous professional evaluation is required. Extensive experiments conducted on three datasets related to  psychological dialogue verify the effectiveness of the proposed method.

\end{abstract}
\section{Introduction}

\begin{figure*}[h]
    \centering
    \includegraphics[width=\linewidth]{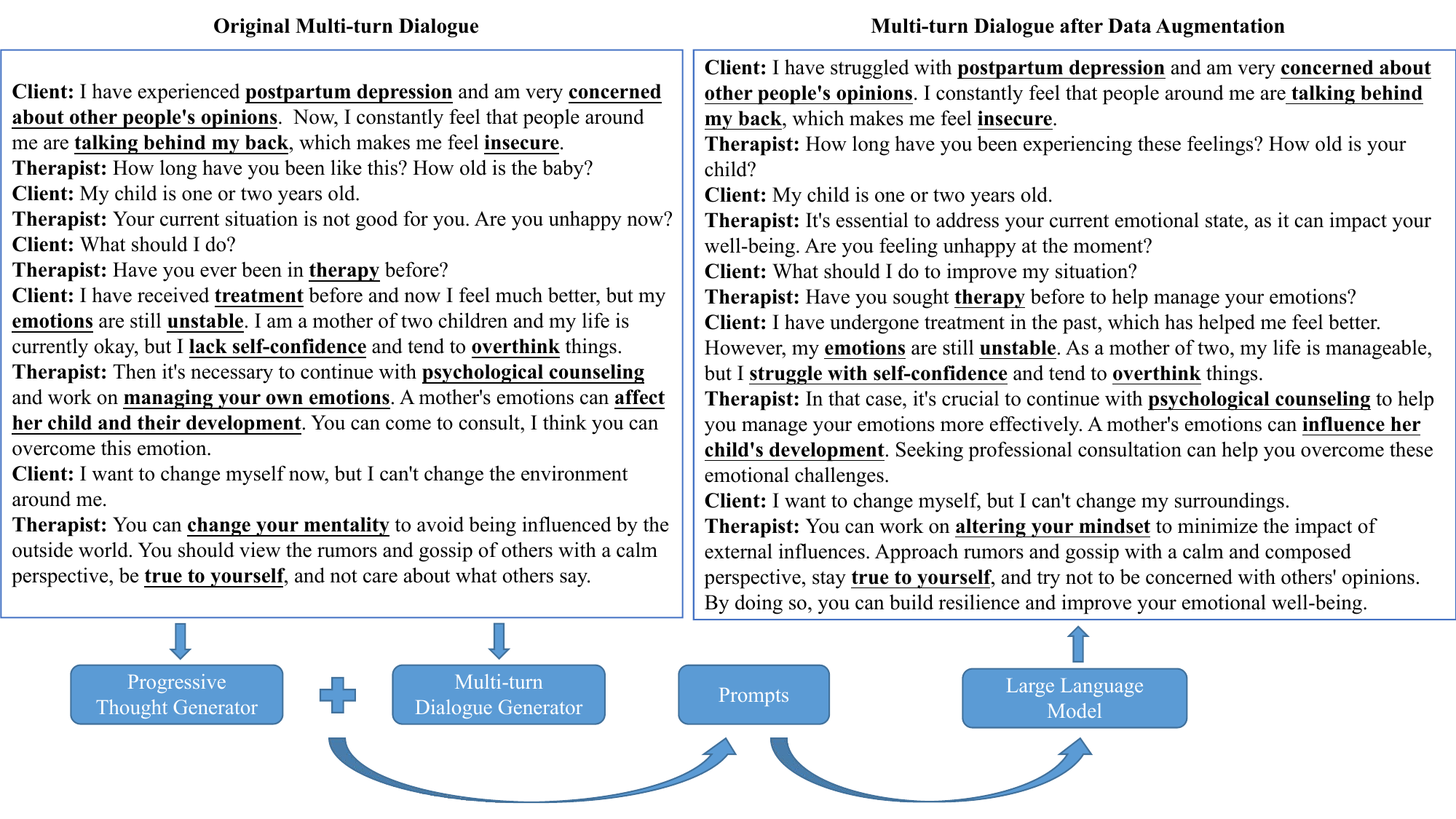}
    \caption{An example of data augmentation for multi-turn psychological dialogue based on the \textbf{KPT} method. The bold and underlined words are keywords, indicating that the keywords did not change significantly after the KPT processing (the contextual semantics did not undergo major changes).}
    \label{fig:example}
\end{figure*}

Data augmentation (DA) of dialogues is a crucial technique in low-resource dialogue tasks.
Lots of contemporary efforts are dedicated to exploring DA techniques to address the issue of insufficient data, that leads to sub-optimal dialogue performance. The majority of these studies \cite{caoetal2022model,guo2022genius,ouetal2022counterfactual} focus on augmenting utterance-level
dialogue rather than the entire dialogue (i.e., multi-turn dialogue). Nevertheless, utterance-level dialogue DA struggles to effectively consider the contextual relationships.
In contrast, multi-turn dialogue aims to generate new conversations based on dialogue history and context, which can naturally address the above issues. 
However, there are few methods for multi-turn dialogue DA, which primarily utilize technologies such as generative adversarial networks \cite{olabiyietal2019multi}, while still facing challenges in handling longer dialogue history and domain-specific dialogue DA. The language representation and dialogue capabilities of smaller models are significantly lower than those of the currently popular LLMs.

The advent of LLMs
has brought numerous exciting developments in DA techniques. Some studies \cite{unknown,dai2023chataug} have demonstrated the remarkable efficacy of LLMs in the DA 
field. 
psychological dialogue is a typical low-resource dialogue task. Previous methods are difficult to incorporate relevant knowledge and specific thinking patterns into psychological dialogue while performing dialogue DA. 

Our goal is to address the data scarcity of psychological dialogue for better performance in low-resource scenarios. Given the rigorous nature of psychological dialogues, considering concrete contexts is preferred in multi-turn data augmentation. We propose a novel \textbf{K}nowledge-driven \textbf{P}rogressive \textbf{T}hought (KPT) prompting method. It includes three components: a progressive thought generator, a psychological knowledge generator, and a multi-turn dialogue generator. The progressive thought generator selects appropriate thoughts from a database to guide multi-turn dialogue generation and prevent semantic deviations. The psychological knowledge generator provides the necessary knowledge, while a penalty evaluation framework ensures dialogue quality. Figure~\ref{fig:example} shows an example of KPT data augmentation for multi-turn psychological dialogue.

In summary, our contributions are as follows: (1) We propose a progressive thought generator, which produces progressive thought across multi-turn dialogue, effectively referencing contextual information and preventing semantic errors in dialogue generation. (2) We introduce a psychological knowledge generator designed to support the creation of psychological knowledge and prompts, enabling better generation of psychological dialogues. (3) By leveraging the powerful capabilities of LLMs in handling context, we select and incorporate knowledge into the dialogue history, which prevents information redundancy and ensures high-quality generation. (4) 
To ensure the precision of multi-turn psychological dialogue generation by LLM, a meticulous professional evaluation is required. Extensive experiments demonstrate the high quality of multi-turn dialogue generated by KPT on three datasets related to psychological dialogue, and the superiority of small models after training based on KPT augmented data. 

\begin{figure*}[h]
    \centering
    \includegraphics[width=\linewidth]{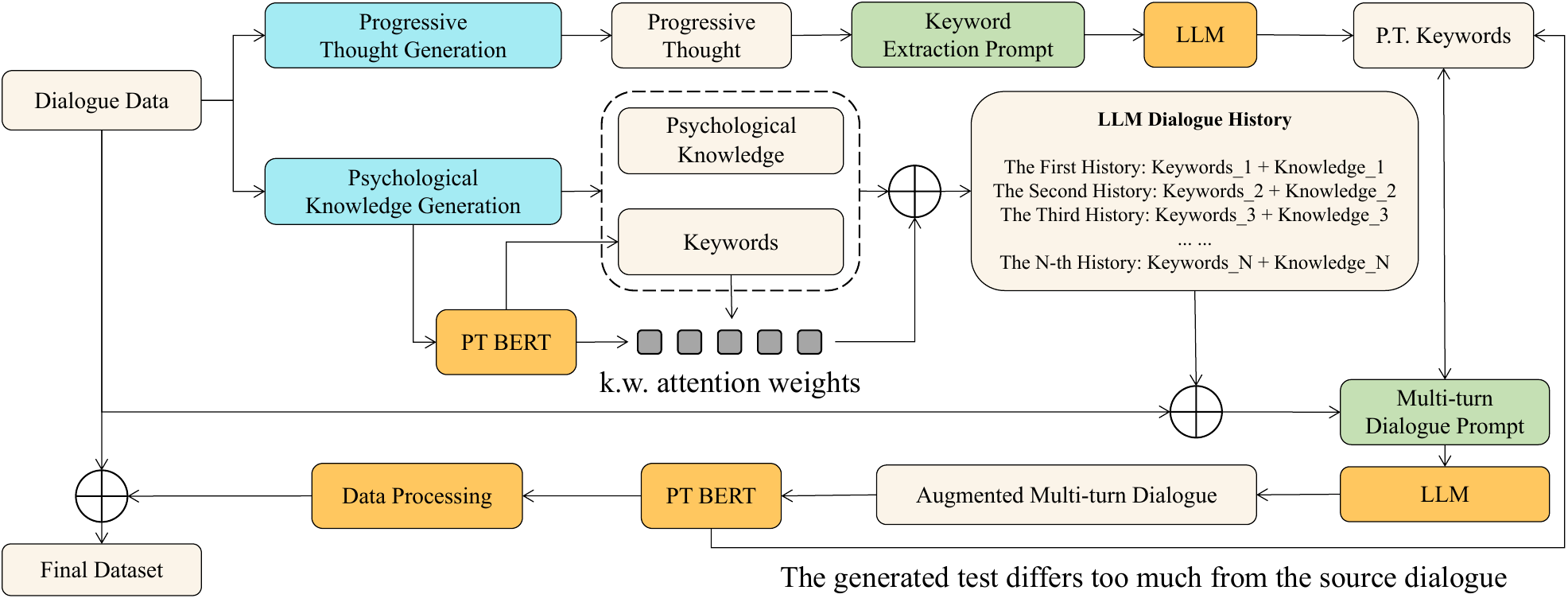}
    \caption{The overview of the proposed KPT.} 
    \label{fig:overview}
\end{figure*}

\section{Method}

The overview of our method is shown in Figure~\ref{fig:overview}. Based on ChatGPT\footnote{\url{https://openai.com/blog/chatgpt}} and BERT \cite{BERTPretraining}, our proposed method KPT consists of three components: (1) Progressive thought generator; (2) Psychological knowledge generator; and (3) Multi-turn dialogue generator. 

\subsection{Progressive Thought Generator}

\begin{figure}[htbp]
    \centering
    \subfigure[Progressive Thought Generator]{
        \includegraphics[width=5.8cm]{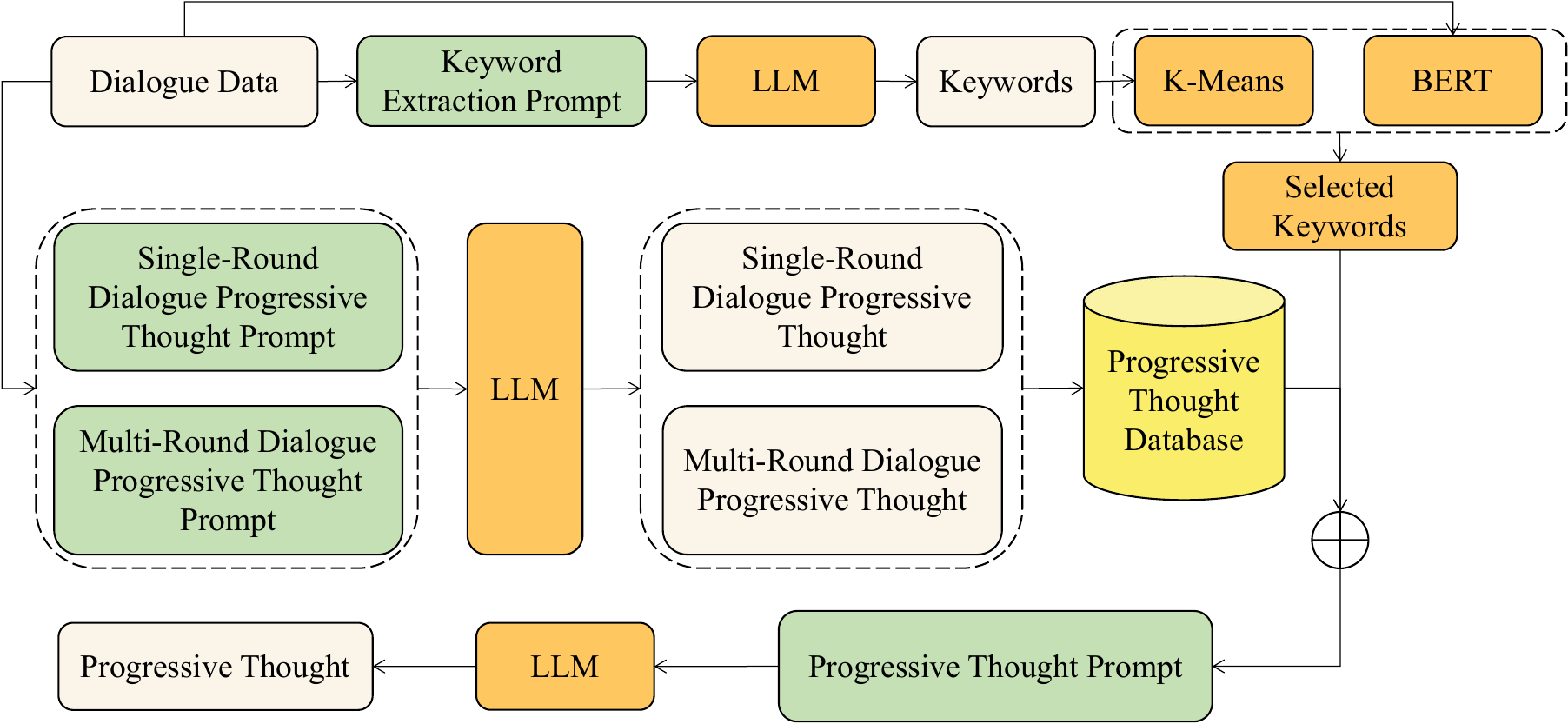}
        \label{thought}
    }
    \hfill
    \subfigure[Psychological Knowledge Generation]{
        \includegraphics[width=5.8cm]{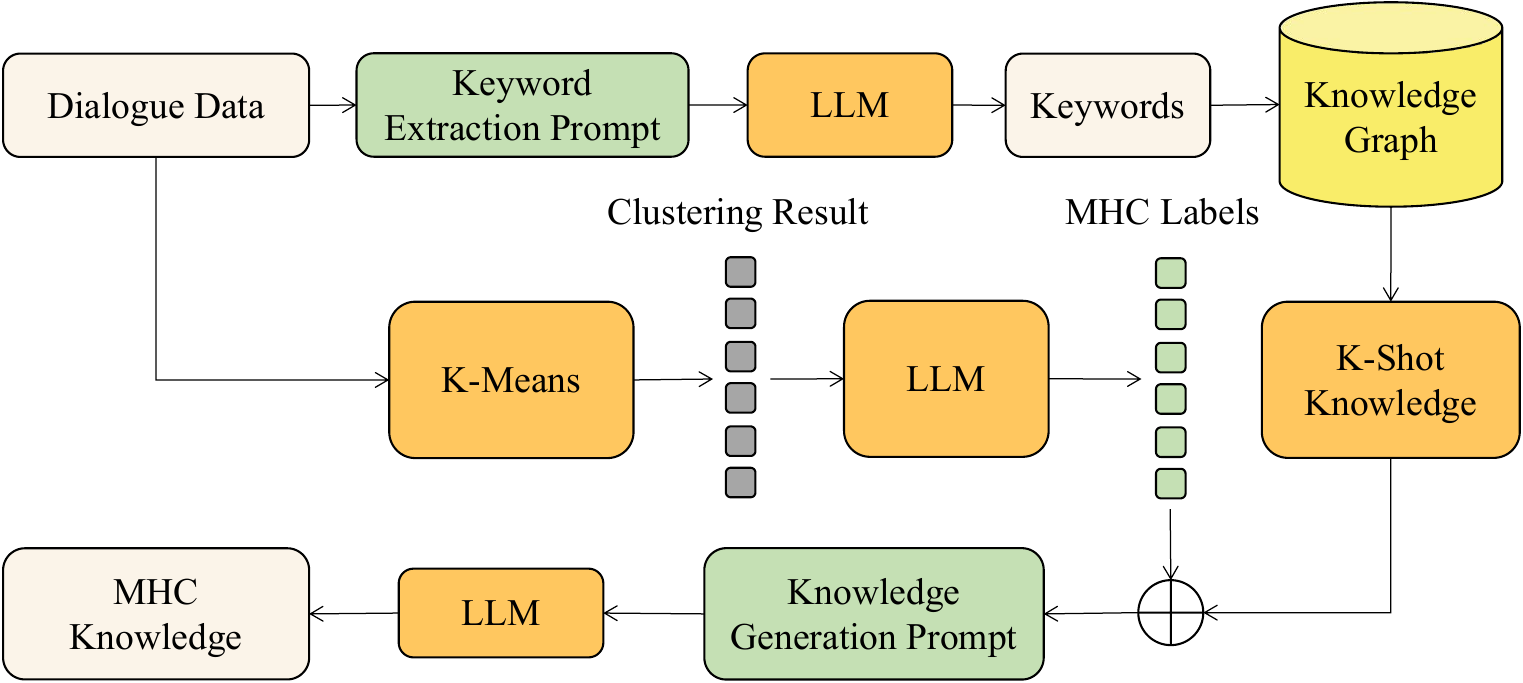}
        \label{knowledge}
    }
    \caption{Progressive thought generator and psychological knowledge generation.}
    \label{fig:both}
\end{figure}

To generate progressive thoughts guiding LLMs in producing multi-turn dialogues with minimal contextual semantic bias, we propose a prompt-based progressive thought generator. As shown in Figure~\ref{thought}, the progressive thought generator comprises multi-turn dialogue keyword selection, progressive thought database construction, and the construction of prompts that facilitate progressive thoughts. 


(1) Multi-turn Dialogue Keyword Selection: We utilize semiotic stimulus (i.e., textual prompt)
as guiding keywords for extraction, which is defined as kw-prompt, to direct LLM in generating dialogue keywords. 
The K-means algorithm is employed to cluster these keywords into a maximum of five categories. To identify the most representative category, a BERT model is used to extract attention weights from the conversation and select keywords from the category with the highest average attention weight as candidate keywords. (2) Progressive Thought Database Construction: Constructing prompts for guiding progressive thought is similar to creating keyword extraction prompts.
We define utterance-level progressive thought prompts and multi-turn progressive thought prompts as ul-prompt and mt-prompt, respectively. These constructed prompts are used to direct the LLM in generating
utterance-level and multi-turn progressive thoughts, respectively. To facilitate subsequent keyword matching to find suitable thoughts, we concatenate all progressive thought and dialogue
to form dialogue-thought combinations (formatted as: utterance of patient + <thought\_prior>  + thought + <thought\_next>  + utterance of therapist).  For multi-turn dialogue, we merge and concatenate the prior dialogue-thought combination with the current one, with at most three rounds of previous dialogue.
Finally, a progressive thought database is formed with multiple progressive thoughts. 
(3) Construction of Prompts that Facilitate Progressive Thought: The candidate keywords are matched with the progressive thought database to select dialogue-thought combinations with the highest keyword occurrence ratio.
Three combinations are used to construct progressive thought prompts if the combination text length is less than one thousand tokens; otherwise, two combinations are used. Then the prompts are employed to guide the LLMs in generating progressive thoughts. 

\subsection{Psychological Knowledge Generator}

To enhance the generation of psychological dialogues, incorporating corresponding knowledge is indispensable. As shown in Figure~\ref{knowledge}, guidance for the LLM in generating psychological knowledge includes three components: k-shot knowledge construction, psychological knowledge label generation, and knowledge generation prompting.


(1) K-shot Knowledge Construction: With the previous progressive thought generator, upon obtaining the keywords of the dialogue, relevant knowledge is retrieved from the knowledge graph (defined as KW-KG).  Since not all keywords can be found within the knowledge graph, only those with corresponding knowledge, counting to k, form data in the format of keyword\_n<knowledge>knowledge\_n, collectively referred as k-shot knowledge:
\begin{equation}
know. = \mathcal{LM}({prompt}_{1}\rm{\textbackslash n}...{prompt}_{k}\rm{\textbackslash n})
\end{equation}
where $know.$~denotes the generated knowledge for the input. Since ``\textbackslash n” is used to separate the prompts, the model will start generating ``\textbackslash n” followed by another random example after finishing the knowledge generation. Hence, we consider the generated sentence before ``\textbackslash n” as $know.$.

(2) Psychological Knowledge Label Generation. We perform k-means on all utterances in the dialogue, clustering them into up to five groups. We take the clustering results and psychological knowledge as input to prompt the LLM, guiding it to transform the clustering results into psychological labels.
(3) Psychological Knowledge Generation Prompting: After the above step, we concatenate the k-shot knowledge with the labeled conversations, forming a prompt (defined as pk-prompt).
This prompt will guide the LLM in generating knowledge related to psychology.

\subsection{Multi-turn Dialogue Generator}

After obtaining the progressive thought and psychological knowledge, the next step is to generate multi-turn dialogue. As shown in Figure~\ref{fig:overview}, the multi-turn dialogue generation component includes a dialogue history organizer and a punishment evaluation framework. (1) Dialogue History Organizer: The output of the psychology knowledge generator consists of psychology knowledge and keywords. By generating dialogue representations with the BERT model, we obtain attention weights for these keywords. Keywords and their corresponding knowledge are sorted based on the weighted average of attention weights, with higher weights placed at the front. The format is shown in Figure~\ref{fig:overview} as LLM Dialogue History. (2) Punishment Evaluation Framework: Following \cite{jiang2023cognitive}, we use the progressive mask method to fine-tune BERT for psychological dialogue. A multi-turn dialogue prompt (mtd-prompt) guides the LLM in generating dialogues. If the similarity between generated and original dialogues exceeds 90\% or falls below 35\%, the dialogues are regenerated. A keyword extraction prompt (pt-ke-prompt) forms a progressive thought keywords database, which randomly selects keywords for guiding dialogue generation. When $n$ is greater than 3, the LLM directly generates dialogues (defined as chatgpt-prompt). Finally, to ensure high data quality, we process the generated dialogues to remove any extra or redundant words and symbols.


\section{Experiments}

\subsection{Implementation Details}

We evaluate our model using three datasets: Chinese Cognitive Stimulation Conversation Dataset (CSConv) \cite{jiang2023cognitive}, Chinese Emotional First Aid Dataset (EFAQA) \cite{efaqa-corpus-zh:petpsychology}, and English Annotated Motivational Interviewing Dataset (AnnoMI) \cite{9746035}. CSConv is a CS-based Chinese conversation dataset that helps the elderly with cognitive impairment maintain or improve their cognitive level. We use the entire 2.6K CS-based conversation dataset as the training and the testing sets. EFAQA is the first open question and answer (QA) corpus in the field of psychological counseling, including 20K pieces of psychological counseling data \cite{efaqa-corpus-zh:petpsychology}. Due to the limitation of API speed, 2.6K dialogue data of EFAQA are randomly selected as training and testing sets. AnnoMI encompasses 133 meticulously transcribed and expertly annotated instances of both high- and low-quality motivational interviewing (MI), a potent therapeutic approach designed to elicit client motivation for fostering positive transformation \cite{9746035}. We use the complete 133 AnnoMI dataset for both the training and testing sets. Since GPT-4 has a three-hour limit of 25 quotas, the LLMs used in our proposed KPT are all ChatGPT. The Chinese BERT models involved in KPT are all fine-tuned or directly used based on bert-base-chinese\footnote{\url{https://huggingface.co/bert-base-chinese}} and the English BERT models is bert-base-multilingual-cased\footnote{\url{https://huggingface.co/bert-base-multilingual-cased}}. We use ownthink knowledge graph\footnote{\url{https://www.ownthink.com/}} as the KW-KG. We implement all models (i.e., PM BERT) in PyTorch \cite{paszke2017automatic} on a single NVIDIA A100 GPU, and train them with AdamW optimization \cite{loshchilov2017decoupled} with a batch size of 4. We vary the learning rate during training following \cite{NIPS2017_3f5ee243}. The training time of KPT (PM BERT) is 3 minutes for about 5 iterations.

\begin{table*}[h]\small
\resizebox{\linewidth}{!}{
\centering
\begin{tabular}{l|c|c|c|c|c|c|c|c|c|c|c}
\toprule
\textbf{Models (Datasets)} & \textbf{B-1} & \textbf{B-2} & \textbf{B-3} & \textbf{B-4} & \textbf{D-1} & \textbf{D-2} & \textbf{Prof.} & \textbf{Eng.} & \textbf{Flu.} & \textbf{Div.} & \textbf{Corr.}\\
\midrule
ChatGPT (CSConv) & 81.57 & 72.03 & 59.64 & 48.15 & 1.42 & 19.27 & 3.24 & 3.87 & 3.56 & 3.33 & 3.45 \\
GPT-4 (CSConv) & 76.44 & 62.57 & 48.39 & 36.17 & 1.50 & 19.74 & 3.56 & 3.98 & 3.78 & 3.80 & 4.01 \\
\textbf{w/o PT (CSConv)} & 74.40 & 62.26 & 50.18 & 39.88 & 1.03 & 15.88 & 3.48 & 3.88 & 3.64 & 
3.62 & 3.48  \\
\textbf{w/o KD (CSConv)} & 66.12 & 53.35 & 41.71 & 32.23 & 0.92 & 17.84 & 3.34 & 3.96 & 3.64 & 
3.68 & 3.86 \\
\textbf{w/o PE (CSConv)} & 80.42 & 70.98 & 58.79 & 45.12 & 1.39 & 17.88 & 3.32 & 4.04 & 3.58 & 
3.78 & 3.65 \\
\textbf{KPT (CSConv)} & 72.48 & 62.12 & 46.76 & 36.32 & 1.41 & 18.98 & \textbf{3.62} & \textbf{4.06} & \textbf{3.86} & \textbf{3.92} & \textbf{4.15} \\
\midrule
ChatGPT (EFAQA) & 86.97 & 76.94 & 64.86 & 52.98 & 0.56 & 19.27 & 3.46 & 3.10 & 3.20 & 
3.86 & 3.48 \\
GPT-4 (EFAQA) & 84.14 & 74.39 & 63.53 & 52.76 & 0.61 & 19.86 & 3.65 & 3.44 & \textbf{3.85} & 4.46 & 3.56 \\
\textbf{w/o PT (EFAQA)} & 83.07 & 73.38 & 62.33 & 51.12 & 0.46 & 11.27 & 3.78 & 3.32 & 3.76 &  3.96 &  3.56 \\
\textbf{w/o KD (EFAQA)} & 78.73 & 68.69 & 57.19 & 46.49 & 0.48 & 12.13 & 3.44 & 3.18 & 3.60 & 
3.94 & 3.62 \\
\textbf{w/o PE (EFAQA)} & 86.12 & 75.84 & 64.02 & 51.90 & 0.57 & 19.30 & 3.68 & 3.25 & 3.76 & 
4.22 & 3.52 \\
\textbf{KPT (EFAQA)} & 80.85 & 69.37 & 59.07 & 48.66 & 0.56 & 19.18 & \textbf{3.92} & \textbf{3.50} & 3.82 & \textbf{4.58} & \textbf{3.75} \\
\midrule
ChatGPT (AnnoMI) & 47.26 & 43.72 & 40.74 & 38.06 & 0.07 & 1.09 & 3.38 & 3.54 & 3.26 & 3.58 & 3.25  \\
GPT-4 (AnnoMI) & 45.79 & 40.95 & 36.00 & 32.04 & 0.02 & 0.47 & 3.56 & 3.86 & 3.46 & 3.98 & 3.70 \\
\textbf{w/o PT (AnnoMI)} & 43.34 & 40.98 & 37.28 & 33.09 & 0.02 & 0.51 & 3.43 & 3.65 & 3.54 & 3.66 & 3.66 \\
\textbf{w/o KD (AnnoMI)} & 36.01 & 33.01 & 31.20 & 30.20 & 0.03 & 0.59 & 3.45 & 3.74 & 3.42 & 3.87 & 3.49\\
\textbf{w/o PE (AnnoMI)} & 47.19 & 44.75 & 41.72 & 37.90 & 0.07 & 1.11 & 3.28 & 3.76 & 3.52 & 3.67 & 3.58 \\
\textbf{KPT (AnnoMI)} & 42.86 & 40.05 & 36.44 & 32.29 & 0.06 & 0.87 & \textbf{3.64} & \textbf{3.98} & \textbf{3.75} & \textbf{4.10} & \textbf{3.72} \\
\bottomrule
\end{tabular}
}%
\caption{For multi-turn dialogue data augmentation, the evaluation includes comparisons between baselines, our proposed KPT, and its individual components (\textbf{ablation experiment}). The first six are automatic evaluation metrics, while the next five are human evaluation metrics. \textbf{Bold text} indicates leading results in human evaluation metrics. Automated evaluation metrics do not show leading results, so they must be analyzed alongside human evaluation metrics.}
\label{TableA}
\end{table*}

\subsection{Baselines}

We conduct extensive experiments to compare the multi-turn dialogue data augmentation of KPT against the following representative baselines: (1) \textbf{ChatGPT}: this is the baseline that provides ChatGPT with a multi-turn dialogue data rewrite prompt, thereby leading ChatGPT to generate multi-turn dialogues. (2) \textbf{GPT-4}: like the prompt provided to ChatGPT, this prompt guides GPT-4 to generate multi-turn dialogue.
In addition, we also conduct a number of experiments (Please refer to \textbf{Supplementary}) to evaluate the performance of dialogue generation on the following representative baselines: (1) \textbf{ChatGLM-fs} \cite{du2022glm}: ChatGLM is a large language model that has been trained on both Chinese and English, allowing it to be bilingual. ChatGLM-fs utilizes 2-shot prompts to generate dialogue responses based on the pre-existing ChatGLM model. (2) \textbf{Alpaca-fs} \cite{chinese-llama-alpaca}: Alpaca is built upon the original LLaMA \cite{touvron2023llama} by incorporating Chinese data for additional pre-training. 2-shot as a prompt instructs Alpaca to generate a response. (3) \textbf{ChatGPT-fs}: ChatGPT-fs utilizes 2-shot prompts to guide the generation of dialogue responses by ChatGPT. (4) \textbf{GPT-4-fs}: GPT-4-fs utilizes 2-shot prompts to guide the generation of dialogue responses by GPT-4. (5) \textbf{GPT2ft} \cite{radford2019language}: GPT2ft refers to the fine-tuning of the GPT-2 model\footnote{\url{https://huggingface.co/uer/gpt2-chinese-cluecorpussmall}} using a training dataset. (6) \textbf{ChatGPT-GPT2ft}: ChatGPT-GPT2ft refers to the fine-tuning of the GPT-2 model using both a multi-turn dialogue dataset generated by ChatGPT and a training dataset. (7) \textbf{GPT-4-GPT2ft}: GPT-4-GPT2ft refers to the fine-tuning of the GPT-2 model using both a multi-turn dialogue dataset generated by GPT-4 and a training dataset.
To better analyze the influence of different components in KPT, we also conduct an ablation study as follows: (1) \textbf{w/o PT}: KPT only uses progressive thought generator. (2) \textbf{w/o KD}: KPT only employs psychological knowledge generator. (3) \textbf{w/o PE}: KPT only utilizes the punishment evaluation framework.

\subsection{Evaluation}
Psychological dialogue evaluation involves two parts: objective analysis using machine learning to assess data augmentation, and subjective analysis with professionals evaluating model-generated dialogues. Feedback is compared across models. A smaller model, trained on these dialogues, undergoes both automatic and human evaluation, ensuring a comprehensive performance evaluation. 


\textbf{(1) Automatic Evaluation.} For multi-turn dialogue data augmentation, following \cite{wuetal2022dg2}, we adopt BLEU \cite{papinenietal2002bleu} as an evaluation metric. In order to calculate the accuracy of different consecutive token amounts, we further employ BLEU-1 (B-1), BLEU-2 (B-2), BLEU-3 (B-3), BLEU-4 (B-4) for evaluation. We also utilize Distinct-1 (D-1) and Distinct-2 (D-2) \cite{lietal2016diversity} as metrics to evaluate the diversity of generated dialogues. For dialogue generation, following \cite{jiang2023cognitive,srivastava2023response}, we employ three evaluation metrics to automatically evaluate the performance of our KPT: BERTScore \cite{bert-score}, Distinct-1 (D-1) and Distinct-2 (D-2) \cite{lietal2016diversity} measure the ratio of unique unigrams / bigrams within the entirety of generated outcomes as a means of representing diversity.

\textbf{(2) Human Evaluation.} To qualitatively analyze model performance in terms of content and psychology, we conduct human evaluations. We randomly select 100 dialogues and their corresponding outputs from our model and baseline models. Due to the excessive number of dialogue turns in the AnnoMI dataset, we extract segments of 4 to 8 turns from each dialogue. Five evaluators, three experienced psychology professionals and two annotators familiar with psychological datasets to evaluate KPT and the baselines. Each metric is rated on five-scale, where 1, 3, and 5 indicate unacceptable, moderate, and excellent performance, respectively.

For data augmentation, we adopt Profession (Prof.), Engagement (Eng.), Fluency (Flu.), Diversity (Div.), Correctness (Corr.) as human evaluation metrics. Prof.~measures the degree of professionalism in the generated dialogues related to psychology. Eng.~is employed to evaluate the conversational context for coherence. Flu.~measures the fluency of the dialogues and the presence of grammatical errors. Div.~measures the diversity of the dialogues. Corr.~measures whether the dialogues contain obvious psychology errors. 

For dialogue generation, following \cite{liuetal2021towards,tu2022misc,jiang2023cognitive}, we adopt Fluency (Flu.), Identication (Ident.), Comforting (Comf.), Suggestion (Sugg.) as metrics. Participants are requested to evaluate KPT and baselines using the following criteria: (1) Fluency: which chatbot's responses were more coherent and comprehensible? (2) Identification: which chatbot delved deeper into your situation and was more effective in pinpointing your issues? (3) Comforting: which chatbot demonstrated greater aptitude in providing comfort? (4) Suggestion: which chatbot offered more valuable recommendations for addressing your concerns? (5) As with the data augmentation evaluation, we also used Profession (Prof.) as an evaluation metric.

\subsection{Results Analysis and Ablation Experiment}

\paragraph{\textbf{Automatic Evaluation of Multi-turn Dialogue Data Augmentation.}} Table~\ref{TableA} shows BLEU values represent dataset similarity, and Distinct values reflect dialogue diversity. ChatGPT and w/o PE top the BLEU scores, signifying their efficacy in data augmentation with minimal dialogue changes. Conversely, w/o KD scores lowest due to deviations caused by excessive knowledge without progressive thought in multi-turn dialogue generation. Progressive thought correction ensures KPT changes align with original dialogue semantics. Overthinking and knowledge interference from w/o PT and w/o KD reduce dialogue diversity in LLM-generated dialogues, resulting in lower Distinct scores. Nevertheless, KPT, with its "Dialogue History Organizer", avoids information redundancy and promotes diversity, thus achieving one of the highest Distinct scores.

\paragraph{\textbf{Human Evaluation of Multi-turn Dialogue Data Augmentation.}} To better evaluate and analyze KPT, we have conducted a professional human evaluation, as shown as Table~\ref{TableA}. In the context of psychological multi-turn dialogue data augmentation which emphasizes professional dialogues in specific domains as well as the engagement of multi-turn dialogue context, the two most important metrics are profession and engagement. On both metrics, KPT outperforms both the baselines and individual components of KPT. As for the other three metrics, on CSConv and AnnoMI, KPT has also shown outstanding performance compared to other baselines. Due to an excess of emojis and text discontinuities in the original EFAQA dialogue dataset, KPT did not achieve the best results on the Fluency metric. However, KPT is very close to SOTA. KPT in the other two indicators also surpasses the baselines.

\paragraph{\textbf{Automatic Evaluation of Dialogue Generation.}} As shown in \textbf{Supplementary}, on CSConv and EFAQA, w/o PT benefits from progressive thought intervention, yielding higher dialogue quality. However, its diversity is lower compared to w/o KD, resulting in a higher BERTScore for w/o PT-GPT2ft but lower for w/o KD-GPT2ft. KPT effectively addresses these shortcomings by generating diverse yet semantically coherent high-quality data, enabling KPT-GPT2ft to achieve SOTA in both BERTScore and Distinct metrics. On AnnoMI, BERTScore of KPT-GPT2ft also reaches SOTA. As Alpaca excels in various dialogue tasks, it generates notably diverse responses when we use psychological dialogues as prompts. Therefore, Except for Alpaca-fs, Distinct of KPT-GPT2ft also reaches SOTA.

\paragraph{\textbf{Human Evaluation of Dialogue Generation}} As shown in \textbf{Supplementary}, human evaluation is essential for assessing dialogue generation quality. Large models often generate irrelevant responses and do not significantly outperform fine-tuned smaller models in human evaluations. Despite having fewer parameters, KPT-GPT2ft produces high-quality responses due to superior training data, achieving SOTA on the first four human evaluation metrics. However, it struggles with complex responses, such as providing suggestions, and does not reach SOTA on the Sugg.~metric for CSConv and EFAQA. Nevertheless, due to fewer suggestions in AnnoMI, KPT-GPT2ft achieves SOTA on the Sugg.~metric.

\paragraph{\textbf{Case Study.}} Referring to the \textbf{Supplementary}, in the multi-turn dialogue data augmentation user case, KPT-augmented data outperforms baselines in dialogue length and sentence diversity with limited semantic deviation. In the dialogue generation case, while KPT-GPT2ft responses are shorter than those from ChatGPT and GPT-4, they're closer to the golden response and offer higher psychological value.

\section{Related Work}


\paragraph{\textbf{Prompting in LLMs.}} Prompting is a method to strategically extract valuable information from LLMs for enhancing performance in various tasks \cite{diao2023active,wang2023describe}. Initially inspired by in-context learning \cite{brown2020language}, prompting has evolved to include techniques like prefix tuning \cite{liliang2021prefix} and p-tuning \cite{liuetal2022p,DBLP:journals/corr/abs-2110-07602}. Prompting methods are categorized into discrete prompts \cite{wallaceetal2019universal,gaoetal2021making,diao2022black} and continuous prompts \cite{qineisner2021learning,liu2021gpt,han2022ptr}. Discrete prompts use template-like tokens, while continuous prompts use optimized vector sequences. Recent advancements in LLMs, such as chain-of-thought (CoT) prompting \cite{wei2022chain,shao2023synthetic,lyu2023faithful}, have shown significant reasoning capabilities.

\paragraph{\textbf{Dialogue Data Augmentation.}} Dialogue data augmentation is divided into utterance-level (text data augmentation) \cite{ouetal2022counterfactual,guo2022genius,caoetal2022model} and multi-turn dialogue data augmentation \cite{olabiyietal2019multi}. For text data augmentation, Easy Data Augmentation (EDA) \cite{weizou2019eda} is highly effective. Deep learning-based methods \cite{chen2023learning}, such as counterfactual \cite{ouetal2022counterfactual} and sketch-based augmentation \cite{guo2022genius}, have shown impressive results. However, utterance-level augmentation can overlook contextual information, leading to biases, making multi-turn dialogue augmentation essential. With the rise of LLMs, dialogue data augmentation techniques based on LLMs show promising results \cite{dai2023chataug,yuan2023llm,liu2023summary}.



\section{Conclusion}

This paper presents knowledge-driven progressive thought prompting for multi-turn dialogues in psychology. KPT combines a thought generator, a psychological knowledge generator, and a dialogue generator to maintain semantic coherence and focus on psychological topics. Evaluations on three datasets confirm KPT's effectiveness. Future work will use augmented datasets to tune LLMs for psychological responses.
%
%
%
%





\bibliographystyle{splncs04}
\bibliography{mybibliography}



\end{document}